\documentclass[sigconf]{acmart}
\AtBeginDocument{%
  }

\copyrightyear{2025}
\acmYear{2025}
\setcopyright{cc}
\setcctype{by}
\acmConference[KDD '25]{Proceedings of the 31st ACM SIGKDD Conference on Knowledge Discovery and Data Mining V.2}{August 3--7, 2025}{Toronto, ON, Canada}
\acmBooktitle{Proceedings of the 31st ACM SIGKDD Conference on Knowledge Discovery and Data Mining V.2 (KDD '25), August 3--7, 2025, Toronto, ON, Canada}
\acmDOI{10.1145/3711896.3737216}
\acmISBN{979-8-4007-1454-2/2025/08}

\usepackage{csquotes}

\usepackage{todonotes}
\usepackage{amsmath,mathtools}
\usepackage{bm,bbm}
\usepackage{color, colortbl}
\usepackage{xcolor}
\usepackage[first=0,last=9]{lcg}
\usepackage[linesnumbered,vlined,ruled]{algorithm2e}
\usepackage{tikz}
\usepackage[capitalize]{cleveref}
\usepackage{threeparttable}
\usepackage{multirow}
\usepackage{booktabs}

\usepackage{orcidlink}
\usepackage{forest}
\SetKwInOut{Input}{input}
\SetKwInOut{Output}{output}

\definecolor{Blue}{HTML}{41B0E4}
\definecolor{level0}{RGB}{59, 150, 30}     
\definecolor{level1}{RGB}{64, 120, 220}    
\definecolor{level2}{RGB}{220, 100, 30}    
\definecolor{level3}{RGB}{218, 153, 56}    
\definecolor{Gray}{gray}{0.9}

\newcommand{\all}{\textsc{All}}
\newcommand{\level}{\textsc{Level}}
\newcommand{\sibling}{\textsc{Sibling}}
\newcommand{\method}{HMCN}
\newcommand{\cl}{HMCL}

\begin{document}

\title[Text-Based Hierarchical Multilabel Classification for Mobile Applications]{Enhancing Text-Based Hierarchical Multilabel Classification for Mobile Applications via Contrastive Learning}

\author{Jiawei Guo} 
\affiliation{%
  \institution{Tencent}
  \city{Shenzhen}
  \state{Guangdong}
  \country{China}
}
\email{javedguo@tencent.com}

\author{Yang Xiao} 
\affiliation{%
  \institution{Xidian University}
  \city{Xi'an}
  \state{Shaanxi}
  \country{China}
}
\email{yxiao@xidian.edu.cn} 

\author{Weipeng Huang} 
\authornote{Corresponding Author.}
\affiliation{%
  \institution{Shenzhen Institute of Information Technology}
  \city{Shenzhen}
  \state{Guangdong}
  \country{China}
}
\email{weipenghuang@sziit.edu.cn}
  
\author{Guangyuan Piao} 
\affiliation{%
  \institution{Independent Researcher}
  \city{Dublin}
  \state{County Dublin}   
  \country{Ireland}
}
\email{parklize@gmail.com}


\begin{abstract}
A hierarchical labeling system for mobile applications (apps) benefits a wide range of downstream businesses that integrate the labeling with their proprietary user data, to improve user modeling. 
Such a label hierarchy can define more granular labels that capture detailed app features beyond the limitations of traditional broad app categories. 
In this paper, we address the problem of hierarchical multilabel classification for apps by using their textual information such as names and descriptions.
We present:
1) \method{} (Hierarchical Multilabel Classification Network) for handling the classification from two perspectives: the first focuses on a multilabel classification without hierarchical constraints, while the second predicts labels sequentially at each hierarchical level considering such constraints;
2) \cl{} (Hierarchical Multilabel Contrastive Learning), a scheme that is capable of learning more distinguishable app representations to enhance the performance of \method{}. 
Empirical results on our Tencent App Store dataset and two public datasets demonstrate that our approach performs well compared with state-of-the-art methods.
The approach has been deployed at Tencent and the multilabel classification outputs for apps have helped a downstream task---credit risk management of users---improve its performance by 10.70\% with regard to the Kolmogorov-Smirnov metric, for over one year. 
\end{abstract}

\ccsdesc[500]{Computing methodologies~Neural networks}
\ccsdesc[500]{Computing methodologies~Supervised learning by classification}

\keywords{Hierarchical Multilabel Classification, Contrastive Learning, App Classification}


\maketitle

\section{Introduction}
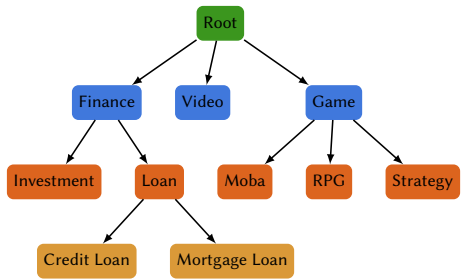
\begin{figure}[t]
\centering
    \scalebox{0.74}{
    \begin{forest}
        for tree={
            draw, thick, rounded corners=3pt,  
            text=black, font=\sffamily,        
            edge={thick, -latex},              
            align=center,
            l sep=0.8cm,                       
            s sep=0.6cm,                       
            if level=0{fill=level0, draw=level0}{
                if level=1{fill=level1, draw=level1}{
                    if level=2{fill=level2, draw=level2}{
                        fill=level3, draw=level3
                    }
                }
            },
        }
        [Root
            [Finance
                [Investment]
                [Loan
                    [Credit Loan]
                    [Mortgage Loan]
                ]
            ]
            [Video]  
            [Game
                [Moba]
                [RPG]
                [Strategy]
            ]
        ]
    \end{forest}
}
\caption{An example of label hierarchy. Moba: multiplayer online battle arena games; RPG: role-playing games.} \label{fg:label_hier}
\end{figure}

Tencent App Store\footnote{\url{https://sj.qq.com/}} is one of the most widely used app stores in China. It has 200 million monthly active users with over 30,000 applications (apps). These apps are organized in a hierarchical structure that can be considered a topic taxonomy. Behind the scenes, apps  are categorized into one or more topics or labels in the taxonomy, which consists of three levels. The top level has broad topics such as \enquote{Finance}, \enquote{Video} or \enquote{Game} and the lower levels have granular topics as shown in~\cref{fg:label_hier}. Organizing items into such a taxonomy is a common practice in industry, and it can be used in many downstream tasks such as providing personalized recommendations for users based on interacted items and associated topics or creating Ad campaigns\footnote{\url{https://developers.google.com/privacy-sandbox/private-advertising/topics}}~\cite{gonccalves2019use,cevahir2016large,huang2020partially,zarrinkalam2020extracting}.
Such a hierarchical structure enables downstream business applications to capture more detailed and comprehensive aspects of apps.
As an example, businesses that run credit risk management can apply the classification results to strengthen their user profiling and modeling, thereby improving their risk management models.

The task of classifying apps into correct topics in the taxonomy can be treated as a hierarchical multilabel classification problem. In our case, an app consists of three types of textual information: 1) the app name, 2) the app description, and 3) the editorial comments summarizing the key functionalities of the app.
Therefore, encoding these textual features into a vector representation or embedding of the app is an important step, which is then followed by building a classification model based on those app embeddings.

Intuitively, we expect the similarity score of app embeddings with similar topics to be higher than that of unrelated ones in the embedding space. This may be observed in the right heatmap in~\cref{fg:sim_emb}.
For example, the similarity between two online video apps (\texttt{IQiYi} and \texttt{Tencent Video}) is much higher than the similarity between unrelated apps like \texttt{Taobao} (a shopping app) and \texttt{Tencent Video}. We can use a straightforward approach such as a pretrained BERT~\cite{devlin2018bert} or one of its variants such as RoFormer~\cite{su2024roformer} to derive the corresponding app embeddings based on the concatenated text of these three types of textual information.
However, as shown in the left heatmap in~\cref{fg:sim_emb}, the similarity scores between app embeddings are not distinguishable using such a straightforward approach, which in turn results in non-optimal classification performance.

Based on the above observations, we propose a Hierarchical Multilabel Classification Network (\method{}) with a pretrained encoder using a Hierarchical Multilabel Contrastive Learning (\cl{}) scheme, for classifying apps in the Tencent App Store.
More specifically, our contributions are highlighted as follows:
\begin{enumerate}
\item 
We propose a Hierarchical Multilabel Contrastive Learning (\cl{}) scheme in Section~\ref{sec:HMCL}, which results in better app embeddings as shown in the right side of \cref{fg:sim_emb}.
\item 
We present the \method{} (Section~\ref{subsec:HMCN}), a hierarchical multilabel classification network incorporating two classification angles: a global one treating all labels in the hierarchy equally, and a local one maintaining different embeddings for predicting the label assignments at different levels.
The experiments in Section~\ref{sec:exp} on our Tencent App Store dataset as well as on two public datasets show that our approach provides the best classification performance on the app dataset, and competitive performance on the public ones.
\item 
The \method{} together with \cl{} has been deployed at Tencent for over one year and the classification results have been used for assessing credit risks of users.
Compared to the previously deployed approach, incorporating app labels from our approach shows 10.70\% improvement on the key evaluation metric, Kolmogorov-Smirnov (KS) value, of the downstream task (Section~\ref{subsec:impact}).
\end{enumerate}
In fact, our solution is more suited to small business teams since it is far less demanding of the computational resources and deployment resources compared to, e.g., large language models~\cite{minaee2024large,zhao2023survey}.

\section{Related Work}
\begin{figure}[t]
	\centering
	\includegraphics[width=0.49\textwidth, trim=0.7cm 0.7cm 0.7cm .7cm, clip=true]{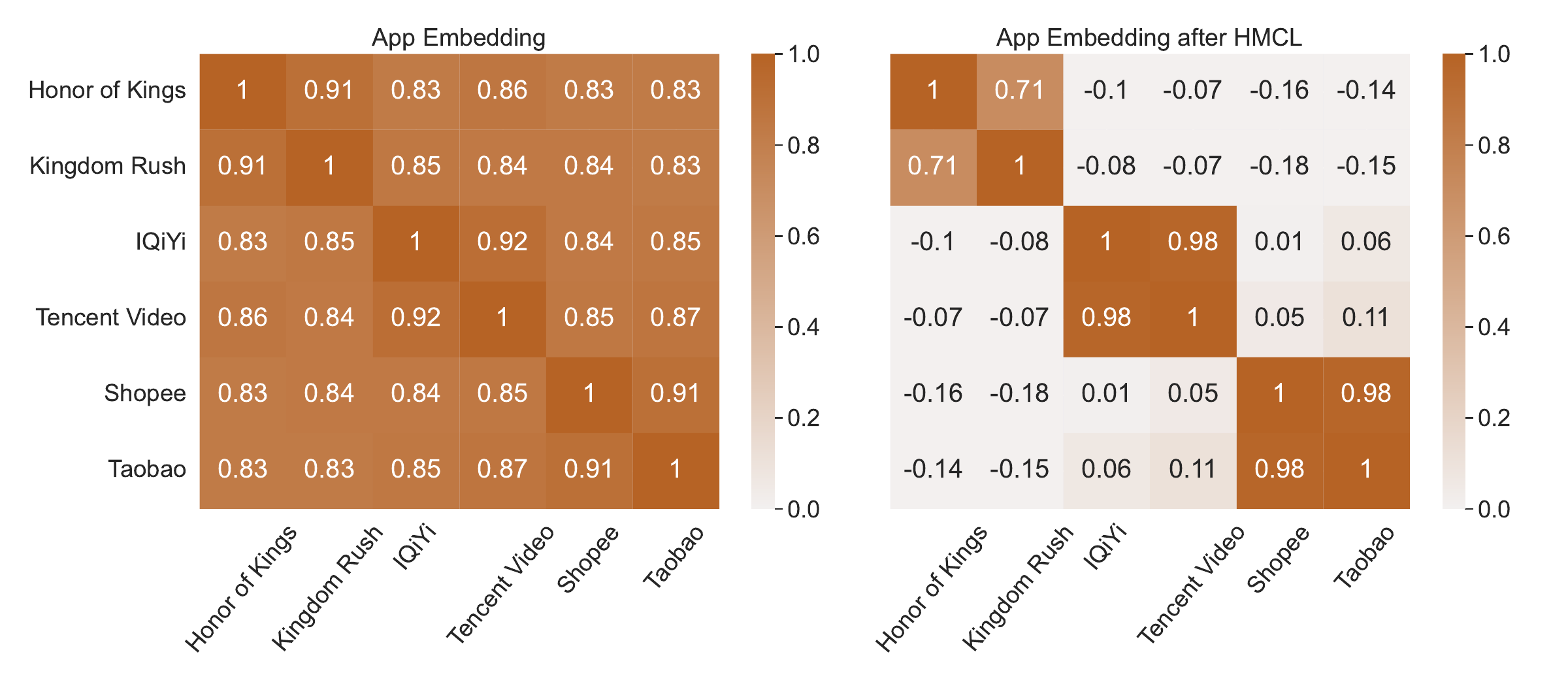}
	\caption{Example of app embedding (cosine) similarities after \cl{} (Hierarchical Multilabel Contrastive Learning)\protect\footnotemark.} 
\label{fg:sim_emb}
\end{figure}

Hierarchical multilabel classification methods can be broadly categorized into global and local approaches~\cite{silla2011survey}.
The global approaches degenerate the hierarchical multilabel classification task into a multilabel classification task~\cite{gopal2013recursive}, which are simple to implement but are often prone to underfitting.
These methods mostly center around developing encoders or decoders that encapsulate the hierarchical constraints~\cite{mao2019hierarchical,you2019attentionxml,rojas2020efficient,zhou2020hierarchy}.
In contrast, local methods pass the information from parent to children, and hence predict labels for each level of the hierarchy from top to bottom~\cite{banerjee2019hierarchical,kowsari2017hdltex,huang2019hierarchical,chen2021hierarchy,shimura2018hft}.
The \method{}~\cite{wehrmann2018hierarchical,xu2021hierarchical} has shown its effectiveness by combining both global and local approaches.
Our work falls into this category and adapts the \method{} to accommodate the multi-field, text-based description of apps.
Motivated by suboptimal app embeddings obtained via a pretrained BERT (or its variant), as illustrated in  \cref{fg:sim_emb}, we introduce hierarchical multilabel contrastive learning for pretraining the text encoder for the \method{}.

\footnotetext{
\texttt{Honor of Kings} and \texttt{Kingdom Rush} are mobile games but with significantly different play styles; \texttt{IQiYi} and \texttt{Tencent Video} are two online video apps; \texttt{Shopee} and \texttt{Taobao} are two online shopping apps.
}

Contrastive learning (CL)~\cite{chopra2005learning,salakhutdinov2007learning,Gutmann2010noise} is a methodology for learning a representation space that allows similar data to get closer together while the pushing dissimilar data further apart.
This approach enhances a model's ability to distinguish between relevant and irrelevant features, ultimately improving its performance on downstream or end-to-end tasks such as classification and clustering.
However, the classical CL has focused primarily on unsupervised learning and multiclass classification~\cite{khosla2020supervised,chen2020simple,he2020momentum,grill2020bootstrap,oord2018representation}.
For multilabel classification in recent studies~\cite{wang2022contrastive,zhang2024multi}, the supervised loss function is modified by introducing weights, derived from the similarity of the label vectors of samples, to adjust the loss for sample pairs.
Wang et al.~\cite{wang2022incorporating} develop a hierarchy encoder and a text encoder to respectively encode hierarchical labels and input text.
It generates a positive peer for a data item by removing the unimportant words from the data item itself.
Zhu et al.~\cite{zhu2024hill} also employ a structural text encoder to encode hierarchical labels and input text.
The work differs from~\cite{wang2022incorporating} in that it projects the label and text vectors into the same space; hence, they define the positive sample pairs as the corresponding label and text vectors.
Although Zhang et al.~\cite{zhang2022use} adopt the term \enquote{hierarchical multilabel contrastive learning}, their approach actually deals with the hierarchies where only one label is assigned to a data point under an active parent label.
In contrast, our case assumes that a data point can be assigned multiple active labels under one active parent label.

Our proposal follows the conventional technical path for CL to focus on the relationships between data samples, rather than the relationships between labels and samples~\cite{wang2022incorporating,zhu2024hill}, taking the hierarchical information into account. Also, our proposed \cl{} occurs during the pretraining phase, prior to training \method{}. It is decoupled from the classification model training and can be used with other classification models.

\section{Model and Implementations}

\begin{figure*}[t]
\centering
\includegraphics[width=1\textwidth]{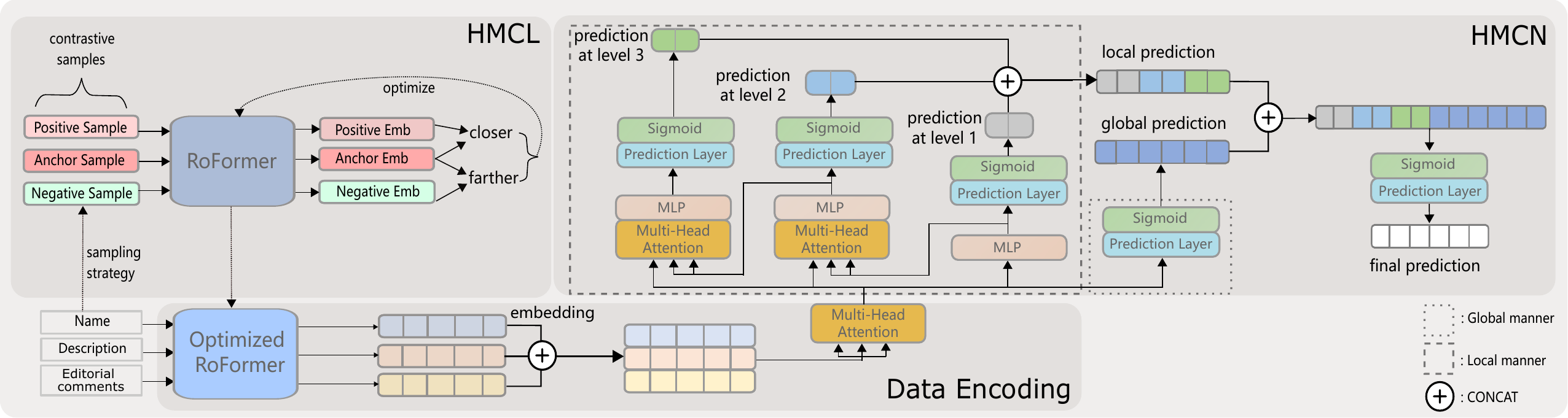}
\caption{The \cl{} + \method{} architecture. 
	In HMCN, the left dotted block refers to the local manner, 
	and the right dotted block refers to global one. The Prediction Layer refers to the last MLP layer for prediction, and {\large \textcircled{\small +}}  indicates a CONCAT operation.} \label{fg:flow}
\end{figure*}

We begin by introducing the necessary notation.
Next, we present our data encoding strategy, followed by a detailed discussion of the \method{} architecture and \cl{}. The overall architecture is illustrated in~\cref{fg:flow}.

\subsection{Notation}\label{subsec:notation}

Using~\cref{fg:label_hier} as a motivating example, we first define a hierarchy  $H$ using the notation of directed graph such that $H = (V, E)$, where $V$  is the label set and $E$ denotes the set of parent-to-child relationships.
Precisely, for a set of $m$ ordered labels, $V = \{v_1, \dots, v_m \}$ and $E = \{ (u, v): u = parent(v), \forall u, v \in V \}$. This clarifies the  top-down paths.
In our task, one parent label can have multiple children while one child label can have only one parent.
We denote the dataset by $\mathcal{D} = \{ \mathbf{X}, \mathbf{Y} \}$ where $\mathbf{X} = \{ x_1, \dots, x_n\}$ represents the features and $\mathbf{Y} = \{\mathbf{y}_1, \dots, \mathbf{y}_n\}$ are the assigned labels to the observations.
Considering any index $i$, $\mathbf{y}_i \in \{0, 1\}^m$ where $1$ indicates that the corresponding label has been assigned to $x_i$ while $0$ refers to the opposite.
Moreover, we denote the assignment of label $v$ for $x_i$ by $y_{iv}$.
We emphasize that there  could possibly be multiple $1$s or all $0$s in any given $\mathbf{y}$.
A label $v$ for any $x$ cannot be $1$ once its parent label $u$ for this $x$ is $0$.
Finally, let us write that $x_i$ has an active label $v$ if $y_{iv}=1$, while $x_i$ has an inactive label $v$ if $y_{iv}=0$.


\subsection{Data Encoding}\label{subsec:data_encoding}
In our scenario, apps are represented by the text data from the App Store. 
We utilize three fields to construct the app embedding: app name, description, and editorial comments.
It follows that we could feed all the information into a text encoder, where we apply RoFormer, to transform the information to embeddings. 
However, concatenating all fields could lead to prohibitively long input text that the encoder has to truncate, resulting in potential information loss.
Also, certain fields might be empty due to practical reasons, and their importance should be down-weighted.
 
To address the above issue, we set a special token to each field with a similar usage of \texttt{[CLS]} in the BERT models~\cite{devlin2018bert}.
This token will be inserted in front of each field and the fields will be independent inputs passed to the encoder.
We extract the embedding for the corresponding token from the three fields and merge them into a 2D embedding.
In particular, we set the tokens \texttt{[N]} for app name, \texttt{[D]} for description, and \texttt{[C]} for editorial comments, respectively.
We input the app name as a sentence $x_N$ into RoFormer and obtain the 2D embeddings $H_N \in \mathbb{R}^{s \times d}$ where $s$ is the length of sequence (without pooling).
As the special token \texttt{[N]} is engineered to always stay in position 0, we can extract an 1D embedding $\mathbf{h}_N \in \mathbb{R}^{d}$  by fetching the first row from $H_N$, such that $\mathbf{h}_N = H_{N}[0] = \mbox{RoFormer}(x_N)[0]$.
Similarly, we achieve the embedding $\mathbf{h}_N, \mathbf{h}_D, \mathbf{h}_C$ for app name, description, and editorial comments respectively by
\begin{align}
	\mathbf{h}_N = H_{N}[0]; \quad \mathbf{h}_D = H_{D}[0]; \quad \mathbf{h}_C = H_{C}[0] \,.
\end{align}
We further obtain the embedding $\mathbf{h}^{*}$ for $x$ by
\begin{align}
	\mathbf{h}^{*} = \mbox{CONCAT}([\mathbf{h}_N, \mathbf{h}_D, \mathbf{h}_C];~~ dim=0) \in \mathbb{R}^{3 \times d}
\end{align}
where $\mbox{CONCAT}(\cdot)$ concatenates the inputs along the $x$-axis when $dim = 0$; otherwise, it concatenates them along the $y$-axis when $dim = 1$.
This formula can then be extended to cases with additional fields.
To focus on the most important features, we apply a self-attention to acquire the embedding at the root level, $\mathbf{h}_0$, by
\begin{align}
\mathbf{h}_0 = \mbox{Encoder}(x) = \mbox{MultiHeadAttn}(\mathbf{h}^{*}, \mathbf{h}^{*}, \mathbf{h}^{*}) \,
\end{align}
where $\mbox{MultiHeadAttn}(\cdot, \cdot, \cdot)$ refers to the multihead attention~\cite{vaswani2017attention}.
The three arguments correspond to the \emph{query}, \emph{key}, and \emph{value} in the attention mechanism.
When all three arguments are the same, it becomes a self-attention computation.

\subsection{Hierarchical Multilabel Classification Network}\label{subsec:HMCN}
We follow the well-known work~\cite{wehrmann2018hierarchical} to split our classification model into two classification manners, namely the global and local manners. 
In the global manner, the constraints of hierarchical multilabel classification are ignored, and all labels are treated equally as in a simple multilabel classification.
The local manner maintains different embeddings for predicting the label assignments at different levels of the hierarchy.
More importantly, the information of the embedding at a higher level (closer to the root node) will be passed on to the embedding at the current level.
This design ensures that the information of the hierarchy can be utilized.
Finally, these two types of predictions are merged to produce the final predictions.

To alleviate the violation of label assignment (i.e., a data point with label $v$ is assigned a value of 1 while its parent $u$ for that data is assigned 0), a path regularization term is added to the loss function.

\subsubsection{Local Manner}
Let $\mathbf{z}$ represent the likelihoods of $\mathbf{y}$ where a single element $z_v = p(y_v = 1 | x)$.
In this manner, the model outputs the estimated likelihoods of the label assignment $\{ \hat{\mathbf{z}}^{(1)}, \dots, \hat{\mathbf{z}}^{(L)} \}$, level by level, where $\hat{\mathbf{z}}^{(\ell)}$ denotes the likelihoods of the predictions at level $\ell$.
Hence, in this local manner, we generate the likelihood estimates $\hat{\mathbf{z}}_{local}$ for $x$ by concatenating the local likelihoods:
\begin{align}
\label{eq:local_pred}
\hat{\mathbf{z}}_{local} = \mbox{CONCAT}([\hat{\mathbf{z}}^{(1)}, \dots, \hat{\mathbf{z}}^{(L)}];~dim=1) \,.
\end{align}

We now describe generation of local predictions at non-root levels.
For any non-root level $\ell$, the embedding $\mathbf{h}_{\ell-1}$ from the last level is gathered to construct the embedding $\mathbf{h}_{\ell}$ along with the meta embedding.
We thus achieve the local embedding by
\begin{align}
\mathbf{h}_{\ell} &=
\begin{cases}
\mbox{MLP}(\mathbf{h}_0) & \ell=1 \\
\mbox{MultiHeadAttn}\left( \mathbf{h}_0, \mathbf{h}_{\ell-1}, \mathbf{h}_{\ell-1} \right ) & \ell=2, \dots, L
\end{cases} \,.
\end{align}
The MLP at the first level can be regarded as a learnable \enquote{prior} for the encoded embedding $\mathbf{h}_0$ and can be thought of as certain information from the root.
For other levels, we acquire the local embedding through cross-attention at the subsequent levels.
We replace the concatenation method used in literature~\cite{wehrmann2018hierarchical} to merge the embeddings by  cross-attentions.
This approach can identify significant features within local embeddings in relation to their parent embeddings, during the information transfer.
In effect, we found that the empirical performance of using multihead cross-attentions and concatenations for information passing is extremely close when testing on our app data.
However, the attention-based approach avoids generating prohibitively long embeddings that may cause memory overflow.

Provided the local embedding, the local prediction is then
\begin{align}
\hat{\mathbf{z}}^{(\ell)} = \mbox{Sigmoid}(\mbox{MLP}_{\ell}(\mathbf{h}_{\ell})), \quad \forall \ell = 1, \dots, L
\end{align}
where the $\mbox{Sigmoid}$ layer is  widely adopted as the multilabel classification layer.
Furthermore, the MLPs for computing the prediction logits also have to be localized for better fitting capacity.
Following that we can construct the local embedding $\hat{\mathbf{z}}_{l}$ using~\cref{eq:local_pred}.

\subsubsection{Global Manner}
The global manner is straightforward to implement.
Let $\mbox{MLP}_g$ denote the global MLP.
We obtain the global prediction $\hat{\mathbf{z}}_{global}$ by
\begin{align}
\label{eq:global_pred}
\hat{\mathbf{z}}_{global} = \mbox{Sigmoid}(\mbox{MLP}_{g}(\mathbf{h}_{0})) \,.
\end{align}

\subsubsection{Prediction Integration}
One may merge the local and global predictions using a weighted mixture.
In our task, the two predictions are combined through an MLP, i.e.,
\begin{align}
\hat{\mathbf{z}} = \mbox{MLP}(\mbox{CONCAT}( [\hat{\mathbf{z}}_{local}, \hat{\mathbf{z}}_{global}] ))
\end{align}
based on~\cref{eq:local_pred,eq:global_pred}.
This approach can also be regarded as a  strategy for acquiring the ensemble of the two predictions.

\subsubsection{Path Regularization}
To mitigate the problem of path violation, we penalize the cases that $p(y_v = 1 | x) > p( y_u = 1 | x)$ where $u$ is the parent label of $v$, for any single $x$.
Given any data point, the likelihood of assigning a value of 1 to the parent label should be at least as high as the likelihood of assigning 1 to any of its child labels.
We adopt a simple hinge loss to define the regularization term $R$ as
\begin{align}
\label{eq:reg}
R(x)
&= \sum_{u, v \in V} \mathbbm{1}\{  u = parent(v)\}\max(0, \hat{z}_v - \hat{z}_u) \,,
\end{align}
such that the sub-term $\hat{z}_v - \hat{z}_u$ can only contribute to the gradient computations when $\hat{z}_v > \hat{z}_u$, i.e., $p(y_v = 1 | x) > p(y_u = 1 | x)$.
In such a case, the sub-term will be minimized.

\subsubsection{Optimization}
We are now able to detail the optimization implementations.
We select the focal loss (FL)~\cite{lin2017focal} as the main loss function, such that
\begin{align*}
\mbox{FL}(\hat{\mathbf{z}}, \mathbf{y})
&= \sum_{v \in V} - \alpha \left[ y_v (1 - \hat{z}_v )^\gamma \log \hat{z}_v + (1-y_v) \hat{z}_v^\gamma \log (1-\hat{z}_v) \right]
\end{align*}
where $\alpha$ and $\gamma$ provide greater flexibility in handling minority labels\footnote{Minority labels refer to the labels that contain only a minor collection of data instances.}, through assigning greater weights to the data points that are hard to learn.
Hence, the final loss function is
\begin{align}
\label{eq:loss}
\mathcal{L}
= \sum_{(x, \mathbf{y}) \in \mathcal{D}} \mbox{FL}(\hat{\mathbf{z}}, \mathbf{y}) + \lambda R(x)
\end{align}
given $\lambda$ a coefficient for weighting the regularization term.
When the value of $\lambda$ is relatively small, the model tends to learn violated paths that favor the statistical characteristics of data over the hierarchical structure~\cite{wehrmann2018hierarchical}.
Conversely, with a relatively large value of $\lambda$, the model tends to assign smaller probabilities to deeper labels, which might affect the convergence process.

\section{Hierarchical Multilabel Contrastive Learning}\label{sec:HMCL}
The \cl{} process is applied prior to training the classification model \method{} to obtain the pretrained starting point.
A good starting point of the model weights could benefit the subsequent training tasks and help to learn a more generalized model~\cite{Erhan20120why}.
One key component to training a successful pretrained model in contrastive learning is constructing effective positive and negative sample pairs
Even though constructing positive and negative sample pairs in multiclass classifications is considered straightforward, the task becomes far more challenging in the multilabel classification scenarios~\cite{wang2022contrastive,dao2021multi,zhang2024multi}, in particular when hierarchical structures are imposed as constraints~\cite{zhang2022use}.
In this work, we propose three negative sampling strategies: \all, \level, and \sibling, which will later be examined in \cref{subsec:neg_sampling_comp}.
In the following discussion, we will detail each of them.

The priority of \cl{} lies in the sampling strategy rather than the conventional contrastive learning loss function design.
The positive sampling procedure is consistent, while the choice of negative sampling strategies will be the key aspect to our empirical success.
Clearly, a pair $(x, x')$ can be a negative pair even if they shared certain labels (but the labels for them are not identical).
They are probably more often regarded as a positive pair if they have more shared labels.
Meanwhile, they could still be a negative pair as long as they have a difference in the label assignment.
With randomness, the sampling approach is equivalent to imposing weights that reflect the degree of label overlap in contrastive pairs, taking into account their hierarchical relationships.
 
\subsection{Positive Sampling}
\label{subsec:neg_samp} 
We adopt the approach of~\cite{zhang2022use}, sampling contrastive data points level by level.
Following this principle, the sampling process for each data point at a given level is performed in a label-wise manner.

Let $V_{i\ell}^+$ denote the positive label set for $x_i$, which contains the active labels for $x_i$ at the $\ell$-th level.
Conversely, let $V_{i\ell}^-$ denote the corresponding negative label set, containing all the inactive labels for $x_i$ at level $\ell$.
While it constructs the label sets given one datum $x$, we also construct sub-sample sets given on every label.
We thus denote the set of sub-samples whose label assignment of label $v$ is active by $\mathcal{X}_v$.
Now, let $\mathcal{X}_{i\ell}^+$ be the positive sample set for $x_i$ at level $\ell$.
We sample one single instance from every positive label of $x_i$, as shown in~\cref{alg:pos_sample}.
%
\begin{algorithm}[t]
\caption{Sampling positive instances for $x_i$ at level $\ell$}
\label{alg:pos_sample}
Initialize $\mathcal{X}_{i \ell}^+ \gets \emptyset$\;
\For{each anchor label $v$ in $V_{i\ell}^+$}{
  Sample $x$ uniformly from $\mathcal{X}_{v}$ and add it to the set $\mathcal{X}_{i \ell}^+$
}
\Return $\mathcal{X}_{i \ell}$
\end{algorithm}

\subsection{Negative Sampling Strategies}
Our solutions focus on constructing effective negative label level-wise that propagate to the samples associated with these labels.
The pair construction process is generally decomposed for each data point and executed level by level.
At each level, negative sampling is performed label by label.
For a given label $v$, referred to as the anchor label in this context, negative labels are sampled relative to $v$.
Assuming that $v$ is at level $\ell$, we focus on constructing the following two sets:
\begin{enumerate}
  \item $V_{\neg v}$ the negative sample set of the anchor label $v$;
  \item $\mathcal{X}^{-}_{i \ell}$ the subset of negative samples of $x_i$ at the $\ell$-th level.
\end{enumerate}
\cref{alg:all} depicts the procedure of this strategy.
\begin{algorithm}[t]
\caption{Negative sampling for $x_i$ at level $\ell$} \label{alg:all}
Initialize $\mathcal{X}_{i \ell}^{-} \gets \emptyset$ \;
\For{each anchor label $v$ in $V_{i \ell}^{+}$}{
  Sample a negative label $u$ from $V_{\neg v}$ \;
  Sample an instance $x$ which has active $u$ but inactive $v$ \;
  Add $x$ to $\mathcal{X}_{i \ell}^{-}$ \; 
}
\Return $\mathcal{X}_{i \ell}^{-} $
\end{algorithm}
For each data point $x_i$, we sample negative samples at every level of the hierarchy.
However, a single data point may have multiple active labels at any given level.
Thus, the algorithm iterates through each active label, treating it as an anchor label $v$. For each anchor label $v$, we construct the corresponding negative label set $V_{\neg v}$ for $v$.
Specifically, with respect to the anchor label $v$, the process involves two key steps:
1) sampling a negative label from $V_{\neg v}$, and
2) sampling an instance associated with this negative label.
Crucially, the sampled instance must not have the anchor label $v$ active, ensuring that the negative sample is distinct and meaningful in the context of CL.
Finally, we construct the level-wise negative sample sets $\{ \mathcal{X}_{i 1}, \dots, \mathcal{X}_{i L} \}$, which are instrumental in computing the final contrastive loss.
Instead of sampling directly from the entire negative sub-space, sampling the negative labels in the first place  ensures that the minority labels can be equally treated and involved in the contrastive learning.

The nature of sampling strategy is to determine the negative sample set that constrains the sampling space, which is in particular directed by $V_{\neg v}$.
Consequently, the methodology used to construct  $V_{\neg v}$ directly shapes the resulting negative sampling strategy.
Next, we explore and analyze three distinct negative sampling strategies: \all{}, \level{}, and \sibling{}.

\subsubsection{Negative Sampling Strategy: \all}
\label{subsec:neg_samp_all} 
To explain this scenario, let us first fix the anchor label for $x_i$ to $v$.
In the \all{} strategy, and the negative label set $V_{\neg v}$ is constructed to include all labels across all levels, excluding the ancestors and successors of $v$. This implies that labels at higher or lower levels in the hierarchy, relative to $v$, are also eligible to be sampled as negative labels for $v$. 
To illustrate the process, we consider the example where the anchor label is set to ``Game''.
In this case, the negative labels must exclude ``Game'' itself, as well as all its ancestors and successors (e.g., ``Game-Moba'', ``Game-Strategy'', etc.).
This strategy promotes broad contrastive comparisons by drawing negative samples from a wide range of labels.
Unfortunately, it turns out that this strategy disregards the hierarchical structure of the labels.

Furthermore, this approach is prone to sampling more negative labels from the lower levels of the hierarchy, as the number of labels grows exponentially when it comes closer to the leaf nodes.
As a result, the negative data samples may be biased towards instances associated with labels that have a large number of leaf descendants, which diminishes the effectiveness of the contrastive comparisons.

\begin{table*}[h!]
	\centering
	\caption{Comparison of contrastive labels generated by three negative sampling strategies, using the label hierarchy in \cref{fg:label_hier} for an example data point with labels \{\enquote{Finance}, \enquote{Finance-Investment}\}.
}
	\label{tb:sampling_strategies}
		\begin{tabular}{p{1cm}p{3cm}p{1.5cm}p{10cm}}
			\toprule
			\textbf{Level} & \textbf{Anchor label} & \textbf{Strategy}& \textbf{Contrastive Label Set} \\
			\midrule
			\multirow{4}{*}{\centering First}
			& \multirow{4}{*}{Finance} & All & \{Video, Game, Game-Moba, Game-RPG, Game-Strategy\} \\
			\cmidrule(l){3-4}
			&  & Level & \{Video, Game\}  \\
			\cmidrule(l){3-4}
			&  & Sibling & \{Video, Game\} \\

			\midrule

			\multirow{5}{*}{Second}
			& \multirow{5}{*}{Finance-Investment} & All & \{Finance, Finance-Loan, Finance-Loan-Credit Loan, Finance-Loan-Mortgage Loan, Video, Game, Game-Moba, Game-RPG, Game-Strategy\} \\
			\cmidrule(l){3-4}
			&  & Level & \{Finance-Loan, Game-Moba, Game-RPG, Game-Strategy\} \\
			\cmidrule(l){3-4}
			&  & Sibling & \{Finance-Loan\} \\
			\bottomrule
		\end{tabular}
\end{table*}

\subsubsection{Negative Sampling Strategy: \level}
This strategy improves \all{} to enhance the negative sample comparison by focusing the negative sample set that stays in the same level of that for the anchor label.
Assume an anchor label for $x_i$ at level $\ell$ is $v$.
Let $V^{(\ell)}$ denote the set of labels at the level $\ell$ of the hierarchy.
The negative set $V_{\neg u}$ is actually the set of labels at the same level except $v$ itself: $V_{\neg v} = V^{(\ell)} \setminus \{v\}$.
The construction of $V_{\neg v}$ leverages hierarchical structure by restricting negative samples to active labels at the same level as the anchor label $v$, explicitly excluding $v $ itself.
Consequently, the labels that are ancestors or successors of $v$ in the hierarchy are inherently eliminated from the negative label pool.
Hence, this strategy can avoid the situation in \all{} strategy, where negative label samples are biased toward those with active labels at lower levels.
Given any anchor label of $x$, we can limit the selection of negative samples to those at the same level, thereby constructing more balanced negative samples given every active label of $x$.

\subsubsection{Negative Sampling Strategy: \sibling}
This approach is inspired by~\cite{zhang2022use}.
The negative sample space for the anchor label $v$ is restricted to its siblings for each level.
Precisely, we have $V_{\neg v}$ to be all the siblings of $v$.
The siblings of $u$ have already excluded $u$ by definition.
In summary, the \sibling{} strategy attempts to better distinguish the data points with respect to the labels sharing the same parent node.
The downside is that it restricts data comparison to non-overlapping negative samples, potentially missing out on certain significant comparisons.

\subsubsection{Comparison of the Negative Sampling Strategies}
\label{par:comparison_of_the_negative_sampling_strategies}
The main distinction between the three strategies is the scope of sampling negative samples based on anchor labels, which further determines different optimization directions.
Noteworthy, for each anchor sample, our approach constructs effective negative labels for every level, propagating to the samples that contain these labels.
We aim to largely avoid cases where negative labels are concentrated in certain labels, as this could hinder the generation of more balanced contrastive samples. 
At a higher level, the \all{} strategy samples negative labels from \emph{all} levels, the \level{} strategy samples from the \emph{current} level of the anchor label, and the \sibling{} strategy samples only from the \emph{sibling} node labels at the anchor label's current level.
The \all{} strategy is prone to sampling anchor labels from the lower levels, which slightly diverges from our goal.
 
Imagine that we would like to sample negative peers for an app with labels \{\enquote{Finance}, \enquote{Finance-Investment}\}.
Using the hierarchy in~\cref{fg:label_hier} as an example, we illustrate the possible sampling results from different negative sampling strategies (see~\cref{tb:sampling_strategies}).
First, we sample the negative labels from the first level.
Let the anchor label at the first level be \enquote{Finance}.
When using the \all{} strategy to select the negative label, our contrastive set is \{\enquote{Video}, \enquote{Game}, \enquote{Game-Moba}, \enquote{Game-RPG}, \enquote{Game-Strategy}\}.
We exclude \enquote{Finance-Loan}, \enquote{Finance-Investment}, \enquote{Finance-Loan-Credit Loan}, \enquote{Finance-Loan-Mortgage Loan} because their first-level labels are also \enquote{Finance}. In other words, they are all the descendant nodes of the anchor label \enquote{Finance}.
Notably, samples from \enquote{Game} or \enquote{Game-XX} share the same first-level label, \enquote{Game}.
Consequently, the probability of sampling the \enquote{Game} label at the first level is four times higher than that of sampling \enquote{Video}, which introduces bias into the optimization process.
The \level{} and \sibling{} strategies address this issue.
The \level{} strategy returns \{\enquote{Video}, \enquote{Game}\} as the set of contrastive labels, since these two labels are both first-level labels.
In the case of the \sibling{} strategy, the contrastive label set is also \{\enquote{Video}, \enquote{Game}\} due to the fact that they share the same parent label \enquote{Root}.

We then sample the negative labels from the second level.
Let the anchor label at the second level be \enquote{Finance-Investment} for our discussion.
When using the \all{} strategy, since the label \enquote{Finance-Investment} has no descendant nodes, the set of contrastive labels is \{\enquote{Finance}, \enquote{Finance-Loan}, \enquote{Finance-Loan-Credit Loan}, \enquote{Finance-Loan-Mortgage Loan}, \enquote{Video}, \enquote{Game}, \enquote{Game-Moba}, \enquote{Game-RPG}, \enquote{Game-Strategy}\}.
In the case of the \level{} strategy, the set of contrastive labels is \{\enquote{Finance-Loan}, \enquote{Game-Moba}, \enquote{Game-RPG}, \enquote{Game-Strategy}\}, since these labels are all second-level labels.
Finally, when using the \sibling{} strategy, the set of contrastive labels contains only one element \enquote{Finance-Loan}, because the label \enquote{Finance-Investment} has only one sibling node.
The \level{} strategy empirically outperforms the \sibling{} strategy since it contrasts each anchor label with a broader distribution of negative sample labels at each level, thereby enhancing the discriminative capabilities.


\subsection{Contrastive Loss}
Let $\mathrm{Proj}(x)$ denote a neural network that employs our encoder to transform input $x$ into $\mathbf{h}$ and a non-linear layer that projects the embedding $\mathbf{h}$ into a subspace, as outlined in the practitioner's guide in~\cite{chen2020simple,grill2020bootstrap,he2020momentum}.
For simplicity, we write $\mathbf{s} = \mathrm{Proj}(x)$.
We emphasize that a normalization layer is always appended to the end of the network, allowing us to perform easy cosine similarity computations.
Most losses in contrastive learning are built on the Softmax function. 
For instance, the InfoNCE~\cite{oord2018representation}, a typical loss and a root for many variants, approximates the probability of correctly identifying the positive sample pairs via a Softmax function over the similarities between the positive and negative pairs. 

However, our scenario is evidently more complex when defining the positive and negative pairs, since there are multiple levels for determining if a sample pair is positive or negative.
What is even more challenging is that the same pair can be regarded as both cases under different anchor labels.
Employing Softmax to model the probabilities of identifying the positive pairs may lead to a cumbersome loss formula and may demand a difficult code implementation.
We therefore propose a more elegant solution, which was found to perform empirically in our task.
Let us denote by $\mathrm{Pr}(x, x^{+})$ and $\mathrm{Pr}(x, x^{-})$ the probabilities of a positive pair $(x, x^{+})$ and of a negative pair $(x, x^{-})$, respectively.
Inspired by multilabel classification, where labels are represented as binary vectors and binary cross-entropy is used as the loss function, we obtain
\begin{align}
\mathrm{Pr}(x, x^{+}) &= \mbox{Sigmoid}(\mathbf{s}^{\top} \mathbf{s}^{+} / \alpha); \\
\mathrm{Pr}(x, x^{-}) &= 1 - \mbox{Sigmoid}(\mathbf{s}^{\top} \mathbf{s}^{-} / \alpha) \,
\end{align}
given $\alpha$ the scaling factor of the input value for Sigmoid.
Recall that $\mathcal{X}_{i \ell}^+$ and $\mathcal{X}_{i \ell}^-$ are respectively the positive and negative sample set for $x_i$ at level $\ell$ of the label hierarchy, generated through the sampling strategies.
Assuming that $I(\mathcal{B})$ returns the indices of data in the batch $\mathcal{B}$, the batch-wise contrastive loss $\mathcal{L}_{cl}(\mathcal{B})$ is
\begin{align}
\mathcal{L}_{cl}(\mathcal{B}) = \frac{1}{|\mathcal{B}|L} \sum_{i \in I(\mathcal{B})} \sum_{\ell=1}^L
\frac{1}{ \left| V_{i \ell}^+ \right|} \Bigg( \sum_{x^+ \in \mathcal{X}_{i \ell}^+} \log \mathrm{Pr}(x_i, x^+) \notag \\
  + \sum_{x^- \in \mathcal{X}_{i \ell}^-} \log \mathrm{Pr}(x_i, x^-) \Bigg)  \,. 
\end{align}
With this loss, one can effectively pretrain the classification model.

\section{Experiments}\label{sec:exp}
In this section, we first present the experimental setup.
Subsequently, we compare the three \cl{} negative sampling strategies, and discuss the experimental findings.
Finally, we discuss the deployment of the \method{} in combination with the \cl{} and its impact on a realworld downstream task.


\subsection{Experimental Settings}\label{subsec:exp_set}
Our primary task was to train a model using the app data\footnote{Part of the public data examples can be found in the Tencent App Store}.
However, to examine the generalization ability of our approach, we also conducted experiments on two public datasets: the RCV1 dataset \cite{lewis2004rcv1} and the WOS dataset \cite{kowsari2017hdltex}.
The RCV1 data contains titles and abstracts (main text).
It is worth noting that WOS is a hierarchical multiclass dataset; thus, there exists only one field of text in the data.
We followed the strategy in \cite{zhou2020hierarchy} to split both datasets.
The statistics of the datasets are summarized in~\cref{tb:datasets}.

\begin{table}[t]
	\centering
	\caption{The statistics of datasets.} \label{tb:datasets}
	\begin{tabular}{cccccc}
		\toprule
		\textbf{Dataset}  & \textbf{Level} & \textbf{Label} & \textbf{Training} & \textbf{Validation} & \textbf{Test} \\
		\hline
		app & 3 & 177 & 30,628 & 3,798 & 14,286 \\ 
		RCV1 & 4 & 103 & 20,833 & 2,316 & 781,265 \\ 
		WOS & 2 & 141 & 30,070 & 7,518 & 9,397 \\
		\bottomrule
	\end{tabular}
\end{table}

Following a general practice~\cite{zhu2024hill,zhou2020hierarchy,aly2019hierarchical}, we adopted micro-F1 and macro-F1 as evaluation metrics.
The micro-F1 computes the F1 score over the entire dataset whereas the macro-F1 is the inter-class average F1 score.
We compared our work with the following models on the RCV1 and WOS datasets:
HiAGM~\cite{zhou2020hierarchy}, HTCInfoMax~\cite{deng2021htcinfomax}, and HiMatch~\cite{chen2021hierarchy}: These models applied a structural encoder to encode hierarchical labels and enhance model performance by matching the semantic similarity between label vectors and text vectors.
HILL~\cite{zhu2024hill} and HGCLR~\cite{wang2022incorporating} are two approaches that employ contrastive learning to augment the representation capabilities of the base text encoders.
Due to the limit of space, we leave our implementation details in the Appendix.

\subsection{Comparing the Negative Sampling Strategies}\label{subsec:neg_sampling_comp}
In this section, we compare the negative sampling strategies for the \cl{}: \all{}, \level{}, and \sibling{}.
\cref{tb:strategies_res} illustrates the comparison of the three strategies on the test sets of the app and the RCV1 datasets, first presenting the performance of a single \method{}, followed by the performance of the \method{} with \cl{} based on each sampling strategy.
Overall, all three strategies outperform the independent \method{} significantly. This indicates that \cl{} enhances the \method{} performance, regardless of which negative sampling strategy is used. 
Of the three negative sampling strategies, \level{} achieves the best performance, followed by \sibling{} and \all{}.
 
Although the \all{} strategy is a close performer to the best-performing \level{} strategy in terms of macro-F1 on the RCV1 dataset, \level{} consistently outperforms it across all results.
This coincides with our analysis in~\cref{subsec:neg_samp} that the contrastive comparisons in \level{} are more effective than that in \all{}.
In \level{}, the selected negative labels for sampling are less biased toward leaf nodes, increasing the chances of picking samples associated exclusively with non-leaf labels.
It enriches the diversity in the contrastive comparisons.
The restrictions on constructing the negative samples in \level{} is the most balanced. 
The \sibling{} strategy outperforms \all{} on all metrics except macro-F1 on the RCV1 dataset. 
Compared with the app dataset, which has evenly distributed labels, RCV1 exhibits a more imbalanced label distribution. We observe that \sibling{} underperforms on under-represented labels in comparison to \all{}, consequently leading to a lower macro-F1 score.
Given the superior performance compared to \all{} and \sibling{}, in the rest of the experiments, we only report the results using the \level{} strategy. That is, \cl{} will be limited to \enquote{\cl{} with the \level{} strategy} unless otherwise noted. 

\begin{table}[t]
  \centering
  \caption{Comparison of the negative sampling strategies.} \label{tb:strategies_res}
  \resizebox{1\columnwidth}{!}{
  \begin{tabular}{lcccc}
    \toprule
        & \multicolumn{2}{c}{\textbf{RCV1}} & \multicolumn{2}{c}{\textbf{app}} \\
        \cmidrule(lr){2-3}  \cmidrule(lr){4-5}
        & \textbf{Micro-F1} & \textbf{Macro-F1} & \textbf{Micro-F1} & \textbf{Macro-F1} \\
    \midrule

    HMCN & $87.52 \pm 0.18$ & $70.39 \pm 0.10$ & $79.67 \pm 0.16$ & $47.79 \pm 0.19$  \\
    \midrule
    \all & $87.13 \pm 0.36$  & $71.14 \pm 0.34$ & $80.16 \pm 0.67$ & $48.02 \pm 1.19$\\
    \textbf{\level} & $\mathbf{87.92 \pm 0.08}$ & $\mathbf{71.36 \pm 0.40}$ & $\mathbf{80.75 \pm 0.05}$ & $\mathbf{48.62 \pm 0.39}$ \\
    \sibling & $87.67 \pm 0.04$ & $70.52 \pm 0.34$ & $80.34 \pm 0.33$ & $48.32 \pm 0.75$ \\
    \bottomrule
  \end{tabular}
  }
\end{table}

\subsection{Results for the App Data}\label{subsec:results_app}
We implemented the BERT and RoFormer model using a global multilabel classification approach.
The two best performing state-of-the-art (SOTA) approaches on the two public datasets, HILL and HGCLR, were also implemented for comparison.
BERT was retained in the implementations, since the HGCLR and HILL are tightly integrated with it.

\begin{table}[b]
  \centering
  \caption{Empirical results on the app dataset.}
  \label{tb:app_exp}
  \begin{tabular}{lcc}
    \toprule
     & \textbf{Micro-F1} & \textbf{Macro-F1} \\
    \midrule
      BERT & $78.28 \pm 0.06$ & $46.59 \pm 0.15$  \\
      RoFormer & $79.19 \pm 0.09$ & $46.57 \pm 0.81$ \\
      \midrule
      HGCLR & 80.62 & 47.78 \\
      HILL & 80.33 & 47.92 \\
    \midrule
      \rowcolor{Blue!20}
	  HMCN & $79.67 \pm 0.16$ & $47.79 \pm 0.19$ \\
      \rowcolor{Blue!20}
      \textbf{HMCN \& HMCL} & $\mathbf{80.75 \pm 0.05}$ & $\mathbf{48.62 \pm 0.39}$ \\
    \bottomrule
  \end{tabular}
\end{table}

\cref{tb:app_exp} shows the results for the app data.
First, we observe that the BERT and RoFormer solutions obtain a close performance despite that RoFormer is better at the micro-F1.
The RoFormer-based HMCN demonstrates a significant improvement on both metrics.
This could be attributed to the local manner of handling classification, and the information transferred from the previous level is helpful.
In addition, we observe that the HGCLR and HILL achieve a close performance.
They both outperformed the \method{} with respect to the micro-F1, and HILL outperforms the \method{} in regard to macro-F1.
However, the \cl{} enables the \method{} to outperform them on both metrics, making our final approach the best-performing one for the data.
This indicates that, the improved quality of app embeddings obtained via \cl{} can effectively enhance the performance of \method. A detailed analysis of the app embedding quality with and without \cl{} regarding uniformity and alignment~\cite{wang2020understanding} can be found in \cref{appendix:emb_quality}.

\subsection{Results for the Public Data}\label{subsec:results}
We report our test set results for the public datasets in~\cref{tb:public_exp}.
As the HGCLR and HILL employ only the field of abstract, we concatenated the title to the abstract when training these two models.
The results of using this modification for these two models both show improvements in the two metrics.
For our own models, the means and standard deviations are presented.
On the RCV1 dataset, the \method{} with \cl{} outperforms all the SOTA solutions.
Interestingly, the performance of the single \method{} (with BERT as the base text encoder) is close to that of the SOTA approaches.
As expected, the \cl{} is able to lift the \method{} to achieve increases in both metrics.
The results imply that the \cl{} effectively helps the instances with active minority labels gain better recognition regarding their representations.

The HILL performs the best for the WOS dataset.
We notice that the WOS dataset contains only one field, namely the main text, whereas the RCV1 dataset contains article titles and main text, and the app data contains more fields.
With only one field in the data, the \method{} might not be able to exploit its full potential since it is designed to cope with data containing multiple fields.
Also, WOS is a hierarchical \emph{multiclass} dataset which might further hinder the \method{} from performing, as the \method{} is designed for a \emph{mutlilabel} scenario.
In particular, the cross-attention for information transfer between levels might be less effective when using only one field.
However, it shows that the \cl{} is consistently capable of enhancing the performance of the \method{}.
This also evidently supports the effectiveness of our contrastive learning procedures.

\begin{table}[t]
  \resizebox{1\columnwidth}{!}{
    \begin{threeparttable}
    \centering
    \caption{Experimental results on the public datasets.}
    \label{tb:public_exp}
    \begin{tabular}{lcccc}
      \toprule
       & \multicolumn{2}{c}{\textbf{RCV1}} & \multicolumn{2}{c}{\textbf{WOS}} \\
       \cmidrule(lr){2-3}  \cmidrule(lr){4-5}
       & \textbf{Micro-F1} & \textbf{Macro-F1} & \textbf{Micro-F1} & \textbf{Macro-F1} \\
      \midrule
        BERT & 86.26 & 67.35 & 86.26 & 80.58 \\
        HiAGM* & 85.58 & 67.93 & 86.04 & 80.19 \\
        HTCInfoMax* & 85.53 & 67.09 & 86.30 & 79.97 \\
        HiMatch* & 86.33 & 68.66 & 86.70 & 81.06 \\
        HGCLR* & 86.49 & 68.31 & 87.11 & 81.20 \\
        HILL* & 87.31 & 70.12 & $\mathbf{87.28}$ & $\mathbf{81.77}$ \\
      \midrule
        HGCLR (title \& abstract) & 87.22 & 69.89 & - & -  \\
        HILL (title \& abstract) & 87.70 & 70.96 & - & - \\
      \midrule
        \rowcolor{Blue!20}
        HMCN & $87.52 \pm 0.18 $ & $70.39 \pm 0.10 $ & $86.45 \pm 0.08$ & $80.91 \pm 0.30$ \\
        \rowcolor{Blue!20}
		{HMCN \& HMCL} & $\mathbf{87.92 \pm 0.08}$ & $\mathbf{71.36 \pm 0.40}$ & $86.90 \pm 0.15$ & $81.07 \pm 0.13$ \\
      \bottomrule
      \end{tabular}
      \begin{tablenotes}
        \item [1] \small{The results for methods marked with * were collected directly from the paper for HILL~\cite{zhu2024hill}}.
      \end{tablenotes}
    \end{threeparttable}
 }
\end{table}

\subsection{Deployment and Impact}\label{subsec:impact}
\subsubsection{App classification}
\label{subsec:hmcn_pipeline}
\begin{figure}[t]
\centering
\includegraphics[width=1\columnwidth]{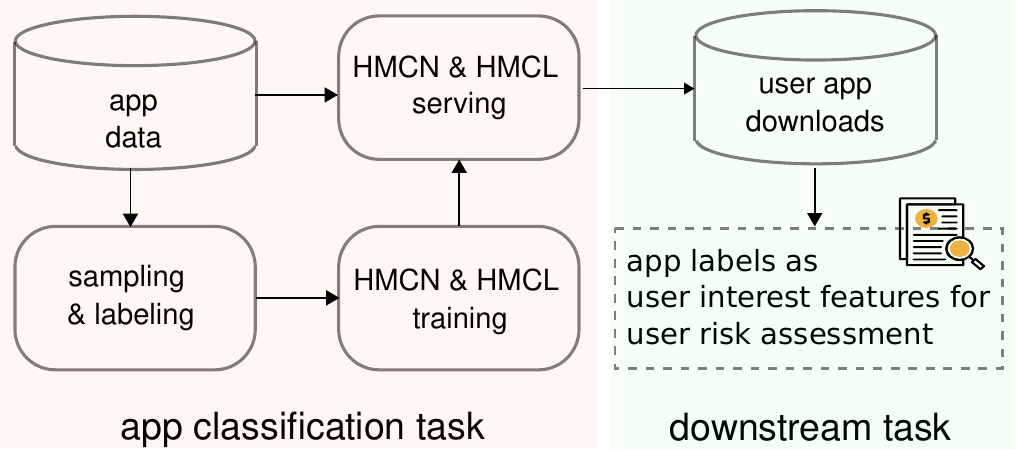}
\caption{The deployment of the \method{} and \cl{}, and the role of its app classifications in the downstream task of user risk assessment. The app labels assigned by the \method{} are used to extract user interests from the app download history of users, which are then merged with additional features for training and inference of the risk assessment model.}\label{fg:deploy}
\end{figure}

As illustrated in illustrated in~\cref{fg:deploy}, we first gather all the app information from the App Store, including the app name, description, editorial comments, etc.
Next, we sample a subset of apps to train and evaluate the model.
This subset will go through several iterations of human labeling to ensure that the resulting ground truth labels are of high quality for training.
After training, the model is deployed to infer labels for all apps.
To account for new and obsolete apps in the App Store, we repeat the inference process monthly to update the labels for all apps.
Moreover, we regularly fine-tune HMCN with newly labeled data to enhance the model.
Given the hyperparameters discussed in the Appendix, the training process of the \cl{} and HMCN would respectively take around 52 hours and 3 hours to complete.
These inferred labels serve as important features that are input into the downstream task.

\subsubsection{Downstream task}
\label{subsec:assessment_pipeline}
We focus our discussion on a particular downstream task where the business objective is to assess the risk of users becoming victims of telecommunications fraud.
Telecommunications fraud refers to the criminal act of defrauding victims into providing confidential information by using false information or disguised identities through communication means such as telephone, text messages, and the internet.
The purpose of this downstream task is to identify in advance the risk of users potentially becoming victims based on user characteristics and to remind relevant organizations to take protective measures. 
The downstream business placeholder could incorporate our classification results to enhance their modeling on user interests.
These user interest features are then merged with other features to train a dedicated machine learning model for assessing user risk. 

To deploy our features online for the downstream task, a dedicated test dataset was prepared and held exclusively by the downstream business team. Features that can improve their evaluation metric beyond a certain threshold on this dataset were approved for integration into their online deployment. Once deployed, the marginal feature contribution would no longer be measured. The positive-to-negative sample ratio was around 1:10, with positive samples referring to the users that could encounter fraud.

Concretely, the effectiveness of user interest features is evaluated using the KS value of the downstream credit risk management model.
The KS value is computed as follows. The downstream model predicts on their user data and generates the scores of these samples being positive. 
We define the cumulative distribution for identifying the positive and negative samples by $\mathrm{CDF}_p(t)$
and $\mathrm{CDF}_n(t)$ respectively, where $t$ is the threshold value and 
\begin{align}
\mathrm{CDF}_p(t)&=\frac{\mathrm{No.\: of\: positive\: samples\: given\: score\:} <\: t}{\mathrm{No.\: of\: positive\: samples}}
\\
\mathrm{CDF}_n(t)&=\frac{\mathrm{No.\: of\: negative\: samples\: given\: score\:} <\: t}{\mathrm{No.\: of\: negative\: samples}}
 \,.
\end{align}
The KS value measures the largest separation between the two CDFs, such that 
\begin{align}
\mathrm{KS} =\mathrm{max}_t \left|\mathrm{CDF}_p(t) - \mathrm{CDF}_n(t)\right| \,.
\end{align}
The higher KS indicates better separation between the two distributions. The business team processed the scores by first sorting them and then dividing them into 11 bins. The threshold values
for each bin were defined by their upper bounds, excluding the first bin.
The metric is used to measure a model's ability to distinguish between positive and negative samples, particularly suited for binary classification problems.
It assesses the model's discriminative power by comparing the maximum difference between the cumulative distribution functions of positive and negative samples~\cite{liu2018comparison}. 
Integrating our app classification results leads to an improvement of $\mathbf{10.7\%}$ with respect to the KS value on the test set. 
It implies that the model can be more accurate ($+10.7\%$) in identifying users who might probably become victims, enabling relevant organizations to take preventive measures in advance.
As a result, app labels classified using \method{} and \cl{} have been successfully deployed online and integrated into the feature set used by the downstream task.
The deployment has been in place for over one year. 


\section{Conclusion} 
In this paper, we proposed a systematic design  for hierarchical multilabel classification of app labels on the Tencent App Store.
Our approach adapts the \method{} to accommodate the multi-field, text-based description of apps.
Additionally, we introduced the \cl{} for pretraining the text encoder, discussed three negative sampling strategies, and examined the effectiveness of leveraging contrastive learning.
With high-quality hierarchical multilabel classification results that are generated by the \method{} with \cl{}, downstream business applications such as user risk assessment can identify the categories of focus and the extent of exploration, which in turn can improve their key performance metrics.
Both offline experiments and 
the 10.70\% performance boost in the downstream task demonstrate the efficacy of the \method{} with the text encoder pretrained through the \cl{}.
Since the textual information such as app names, descriptions, and editorial comments is generally available in other app stores, we believe our approach can be applied to many other app stores as well.
For future work, additional downstream business applications will be explored to leverage the hierarchical multi-labels of apps.

\begin{acks}
We are grateful to the anonymous reviewers for the extremely constructive feedback, which has helped significantly improve this paper.

This project was initiated when Weipeng Huang was with Tencent.
Weipeng Huang has been partly supported by Doctoral Research Initiation Program of Shenzhen Institute of Information Technology (Grant SZIIT2024KJ001) and Guangdong Research Center for Intelligent Computing and Systems (Grant PT2024C001).
\end{acks}

\bibliographystyle{ACM-Reference-Format}
\bibliography{references}

\appendix


\section{Implementation Details}
We now specify the implementation of the contrastive learning.
As discussed in~\cref{sec:HMCL}, we sample positive and negative instances for anchor samples at each level. 
To ensure sufficient contrast among samples, we repeated this process multiple times.
For the app dataset, we respectively performed this sampling process 10, 20, and 50 times for level 1, 2, and 3.
For the RCV1 dataset, we respectively performed the sampling process 10, 20, and 50 times for level 1, 2, and 3.
Since there is only one label at level 4, we would not conduct sampling at this level.
At last, for the WOS dataset, we respectively repeated the sampling process 5 and 20 times, for level 1 and 2.

We learn that the number of samples per label, at higher hierarchical levels, is less than that at lower levels.
Thus, the sampling multiples are correspondingly higher.
After repeating the sampling for each anchor sample multiple times, we obtain a sufficient number of contrastive samples.
We used the Adam optimizer with a mini-batch size of 8 and a learning rate of $\text{1e-5}$ for all configurations.
The learning rate decayed based on the number of training batches, reducing by a factor of 0.8 every 4,000 batches.
Additionally, we set the scaling parameter $\alpha$ in the contrastive loss function to 0.1.
As mentioned above, we repeated the same sampling strategy for each batch multiple times, we indeed increased the number of comparisons.
Consequently, we observed that 1 epoch for the \cl{} is always sufficient to perform.

With regard to the base text encoder, we applied RoFormer\footnote{\url{https://huggingface.co/junnyu/roformer_chinese_base}} for the app data, while BERT\footnote{\url{https://huggingface.co/google-bert/bert-base-cased}} was employed for both RCV1 and WOS datasets.
The text encoder trained through the~\cl{} was used in the followed classification tasks.
We applied the Adam optimizer with a batch size of 8 for all datasets.
For all classification settings, we trained the model for 20 epochs and decayed the learning rate by a factor of 0.8 every two epochs.
Regarding the app data, we set the initial learning rate to $\text{5e-3}$.
For the RCV1 and WOS datasets, we set the initial learning rate to $\text{1e-4}$.
The classification experiments for the public datasets were repeated 5 times under each configuration.

The hyperparameters for the focal loss were fixed to the default values $\alpha=0.25, \gamma=2$.
The scaling factor for the path regularization $\lambda$ was fixed to a simple choice of $1$ under all configurations.
When gauging the labels from the prediction logits, we applied the threshold of $0.5$ to decide if a label $v$ is considered active for $x_i$.
That is, given any data index $i$ and label $v$, we defined
 \[
 y_{i v} = \begin{cases}
   1 & z_{iv} \ge 0.5 \\
   0 & otherwise
 \end{cases} \,.
 \]
 
Our experiments were conducted on a server equipped with the Xeon(R) Platinum 8372HC CPU with 3.40GHz, 8 NVIDIA A10 Tensor Core GPUs, and 360GB of RAM.
The floating-point precision was set to BF16 (Brain Floating Point 16 bits).

\section{Uniformity and Alignment of App Embeddings with and without the HMCL}
\label{appendix:emb_quality}
Here, we examine two important properties, \emph{uniformity} and \emph{alignment}, to assess  the quality of app embeddings without and with \cl{}.
Uniformity measures how well the embeddings are spread out over the representation space, while alignment measures how close embeddings of positive pairs are in the  representation space. They are formally defined as follows.

The \emph{uniformity} property favors embeddings that are roughly uniformly distributed on the unit hypersphere, preserving as much information of the data as possible~\cite{wang2020understanding}.
For randomly sampled pairs $(x, \mathbf{y})$ with normalized embeddings $\mathbf{s}_x, \mathbf{s}_y $, the uniformity loss is formulated as:
\begin{equation}
    \mathcal{L}_{\text{uniform}} = \log \mathbb{E}_{(x, \mathbf{y}) \sim p_{\text{data}}} \left[ \exp \left \{\tau \left(\mathbf{s}_x^\top \mathbf{s}_y - 1 \right) \right\} \right]
\end{equation}
where $\tau > 0$ is the temperature hyperparameter, $p_{\text{data}}$ is the data distribution, and $\mathbf{s}_x^\top \mathbf{s}_y - 1 = -(1 - \mathbf{s}_x^\top \mathbf{s}_y)$ represents the negative cosine distance between $\mathbf{s}_x$ and $\mathbf{s}_y$.

The \emph{alignment} property prefers that two samples forming a positive pair should have embeddings that are close each other, and thus be (mostly)
invariant to irrelevant noise factors~\cite{wang2020understanding}.
For positive pairs $(x, x^+)$ with normalized embeddings $\mathbf{s}, \mathbf{s}^+$, the alignment loss is defined as:
\begin{equation}
	\mathcal{L}_{\text{align}} = \sum_{l=1}^L \mathbb{E}_{(x, x^+) \sim p_{\text{pos}}^{(l)}} \left[ 1 - \mathbf{s}^\top \mathbf{s}^+ \right]
\end{equation}
where $\mathbf{s}^\top \mathbf{s}^+$ represents cosine similarity and $1 - \mathbf{s}^\top \mathbf{s}^+$ denotes the corresponding cosine distance.
Apart from that, $p_{\text{pos}}^{(l)}$ denotes the distribution of positive pairs at level $l$, where the two samples $x$ and $x^+$ share at least one common label at that level.

\begin{figure}
\centering
\includegraphics[width=0.45\textwidth]{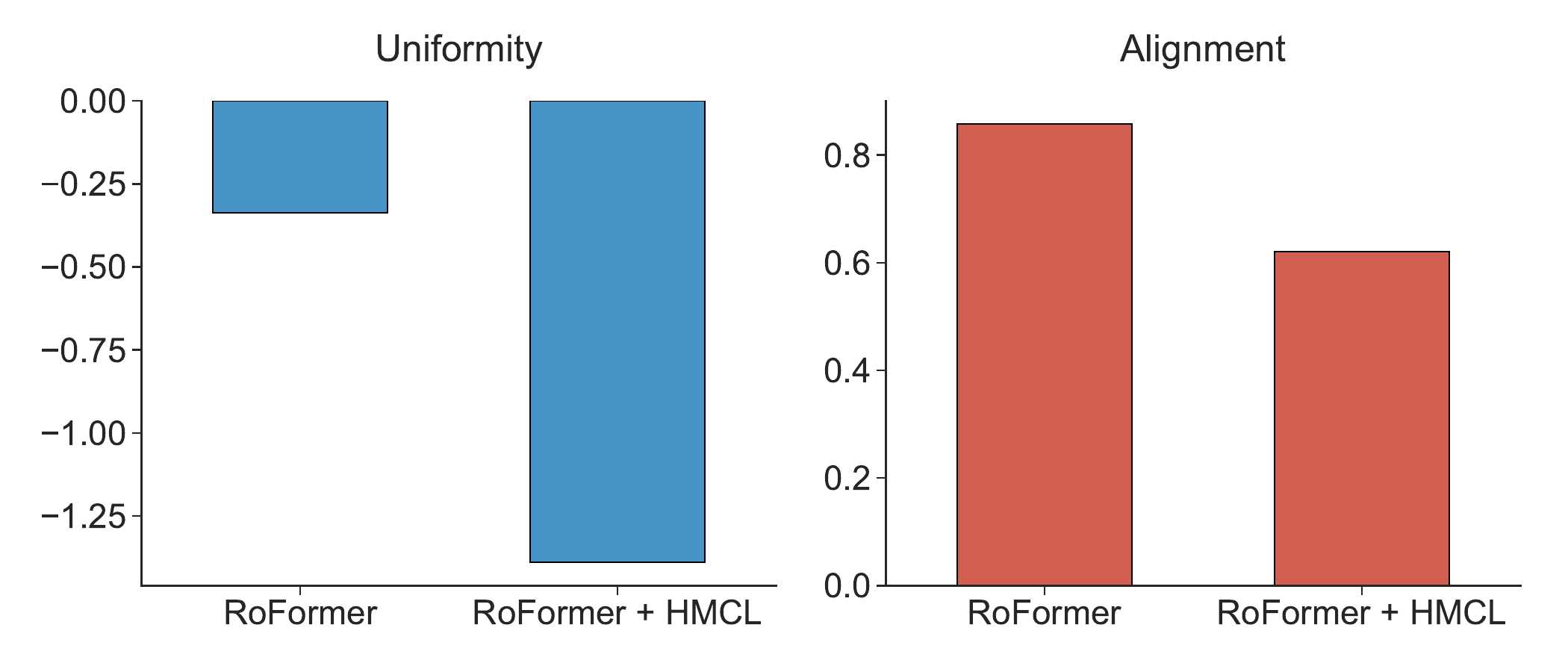}
\caption{Improvement of app embedding quality with \cl{} (lower values indicate better performance for both metrics)} \label{fg:alignment_uniformity}
\end{figure}

For both metrics, lower values indicate better performance. As shown in \cref{fg:alignment_uniformity}, uniformity improves (decreases from -0.337 to -1.390), and alignment also improves (decreases from 0.858 to 0.602), indicating enhanced app embedding quality with \cl{}.

%

\end{document}